%% file: Template_ISBI_latex.tex
\documentclass{article}
\usepackage{spconf,amsmath,graphicx}
 \usepackage{titlesec}
\usepackage{pifont}% http://ctan.org/pkg/pifont
\newcommand{\cmark}{\ding{51}}%
\newcommand{\xmark}{\ding{55}}%
\usepackage{amsmath, amssymb, amsfonts}
\usepackage{booktabs}
\usepackage[framemethod=tikz]{mdframed}
\usepackage[most]{tcolorbox}
\newtcolorbox{softgraybox}{
  colback=black!10, colframe=black!25, boxrule=0.8pt, arc=1mm,
  left=2pt,right=2pt,top=2pt,bottom=2pt
}
\usepackage{algorithm}
\usepackage{algorithmic}
\usepackage{subcaption}   % or \usepackage{subfig}

\usepackage{amsmath, amssymb, amsfonts}
\usepackage{booktabs}
\usepackage{multirow}
\usepackage{tabularx}

\usepackage{enumitem}
\setlist{nosep, leftmargin=14pt}

\usepackage{mwe} % to get dummy images

\title{ProGiDiff: Prompt-Guided Diffusion-Based Medical Image Segmentation}
\name{
  \begin{tabular}{c}
    Yuan Lin$^{1}$, Murong Xu$^{2}$, Marc H\"olle$^{1}$, Chinmay Prabhakar$^{2}$,\\
    Andreas Maier$^{1}$, Vasileios Belagiannis$^{1}$, Bjoern Menze$^{2}$, Suprosanna Shit$^{2}$
  \end{tabular}
}
\address{
  $^{1}$Friedrich-Alexander-Universität Erlangen-Nürnberg, Germany\\
  $^{2}$University of Zurich, Switzerland\\
  yuan.lin@fau.de
}

\begin{document}
%\ninept
%
\maketitle
\begin{abstract}
Widely adopted medical image segmentation methods, although efficient, are primarily deterministic and remain poorly amenable to natural language prompts. Thus, they lack the capability to estimate multiple proposals, human interaction, and cross-modality adaptation. Recently, text-to-image diffusion models have shown potential to bridge the gap. 
However, training them from scratch requires a large dataset—a limitation for medical image segmentation. Furthermore, they are often limited to binary segmentation and cannot be conditioned on a natural language prompt. 
To this end, we propose a novel framework called ProGiDiff that leverages existing image generation models for medical image segmentation purposes. Specifically, we propose a ControlNet-style conditioning mechanism with a custom encoder, suitable for image conditioning, to steer a pre-trained diffusion model to output segmentation masks. It naturally extends to a multi-class setting simply by prompting the target organ. Our experiment on organ segmentation from CT images demonstrates strong performance compared to previous methods and could greatly benefit from an expert-in-the-loop setting to leverage multiple proposals. Importantly, we demonstrate that the learned conditioning mechanism can be easily transferred through low-rank, few-shot adaptation to segment MR images.
% Widely adopted medical image segmentation methods, although efficient, are primarily deterministic. Thus, they lack the capability to estimate uncertainty and a transferable prior without incurring additional computation via ensemble or probabilistic modelling, making them suboptimal. Recently, diffusion models have attempted to bridge the gap; however, training them from scratch requires a large dataset—a limitation for medical image segmentation. Furthermore, they are often limited to binary segmentation and cannot be conditioned on a natural language prompt—a feature known to assist cross-modality adaptation. To this end, we propose a novel framework called ProGiDiff that leverages existing image generation models for medical image segmentation purposes. Specifically, we propose a ControlNet-style conditioning mechanism with a custom encoder, suitable for image conditioning, to steer a pre-trained diffusion model to output segmentation. It naturally extends to a multi-class setting simply by prompting the target organ. Our experiment on organ segmentation from CT images demonstrates strong performance compared to previous methods and could greatly benefit from an expert-in-the-loop setting to leverage multiple proposals. Importantly, we demonstrate that the learned conditioning mechanism can be easily transferred through low-rank, few-shot adaptation to segment MR images.
\end{abstract}
\input{section/01_intro}
\input{section/02_rel_lit}

\input{section/03_method}
\input{section/04_exp}

\section{Conclusion}

In this paper, we propose a prompt-guided diffusion-based model, named ProGiDiff, for multi-label abdominal organ segmentation. 
We explored the use of Stable Diffusion with ControlNet to perform segmentation tasks in a controllable and interpretable manner. Our approach also enables few-shot segmentation for MRI by applying LoRA-based fine-tuning to ControlNet. Experiments on the BTCV and CHAOS datasets demonstrate strong performance and generalization under limited training data. The generative nature and language-promoting feature of our model are suitable for an expert-in-the-loop system, such as a routine clinical pipeline.

% \input{section/07_supp}

% \section{Acknowledgments}
\clearpage
\bibliographystyle{IEEEbib}
\bibliography{strings,refs}

\end{document}

%% file: section/01_intro.tex
\section{Introduction}
\label{sec:intro}

Medical image segmentation is a fundamental task in computer-aided diagnosis. 
% It aims to simplify the complexity of medical images by decomposing them into multiple meaningful anatomical regions~\cite{kazerouniDiffusionModelsMedical2023}. 
Accurate segmentation allows for quantitative assessment of treatment outcomes and supports efficient clinical decision-making~\cite{xu2025cads}.
%,hanDeepSemisupervisedLearning2024
% Consequently, not just high‑precision segmentation, but also identifying uncertainty is of paramount importance for its reliable and safe real-world deployment. 
% Since the introduction of U-Net~\cite{ronneberger2015u}, a wide range of segmentation methods aim for performance optimization~\cite{huang2020unet, isensee2024nnu, cai2022ma} including more-recent transformer adaptation \cite{chen2021transunet, cao2022swin, ji2021multi}. 
% However, traditional U-Nets, constrained by their limited receptive fields, struggle to understand complex vision information when dealing with intricate and diverse structures. More recently, researchers have adapted transformer architectures~\cite{Vaswani2017AttentionIA} for medical image segmentation~\cite{chen2021transunet, cao2022swin, ji2021multi} by modeling the long-distance dependence between pixels to obtain global context information through self-attention mechanisms~\cite{pu2024advantages}, achieving improved segmentation accuracy. 
However, state-of-the-art (SOTA) approaches~\cite{isensee2021nnu, 3Dtransunet, cao2022swin} are inherently deterministic and provide no natural-language controllability: clinicians can not explicitly specify the target anatomy. 
Recently, pretrained generative models such as Stable Diffusion~\cite{rombach2021highresolution}, trained on massive natural image corpora, offer strong visual priors and support text conditioning, which makes them attractive for prompt-guided segmentation and for generating multiple plausible proposals. 

A key challenge is that existing pre-trained generative models target natural-image synthesize rather than segmentation masks with anatomical structures.  And conditioning medical image segmentation on natural language prompts is a challenging task due to the difficulty in achieving alignment between textual and visual modalities~\cite{zhou2020unified}. 
To address these challenges, we propose a novel framework for \emph{prompt-based segmentation leveraging a text-conditioned image diffusion model}. Our approach (i) serves as an effective solution for segmenting an arbitrary number of classes using the same diffusion model. (ii) Encodes semantic prior information into the model and (iii) provides language-driven interpretability. However, to truly unlock the capabilities of our approach, it is essential that our method can be easily adapted to different imaging modalities. 
% This leads us to our third research question:
% \begin{softgraybox}
% Q3. How can we easily adapt learned semantic prior to a different modality?
% \end{softgraybox}
We argue that a prompt-driven semantic prior can bridge cross-modal information transfer (e.g., CT to MR). 
% However, the model needs to be adapted to handle the substantial variation between imaging modalities. 
To this end, we propose a low-rank adaptation of the controller and fine-tuning of the image encoder. We validate our approach through experiments demonstrating robust generalization across modalities. 
% In summary, our contribution is as follows.

% \begin{mdframed}[hidealllines=true,backgroundcolor=blue!5]
\paragraph{Contributions:} 
In this work, we propose a novel approach called ProGiDiff to leverage a pretrained image diffusion model for the prompt-guided medical image segmentation task. Our method can utilize language prompts for multi-class segmentation, addressing a significant limitation of existing methods. 
On CT abdominal organ segmentation, ProGiDiff attains competitive performance compared to existing diffusion models and exhibits excellent few-shot domain adaptation when applied to an MR dataset. 
% transfers to MR with few-shot adaptation. In addition, it can generate multiple segmentation proposals, facilitating expert-in-the-loop refinement.
% Not only do our experiments on organ segmentation from CT images achieve encouraging results compared to existing diffusion models, but more importantly, our method exhibits excellent few-shot domain adaptation when applied to an MR dataset. 
Importantly, we demonstrate that multiple segmentation proposals are well-suited for an expert-in-the-loop setup. 
% Furthermore, we qualitatively reveal that our proposed method captures object-centric uncertainty more effectively than the previous pixel-centric one of the diffusion-based method.
% \end{mdframed}

%% file: section/02_rel_lit.tex
\section{Related Work}
\label{sec:rel_lit}

% \textbf{Diffusion-based Medical Image Segmentation:}
Latent diffusion models have recently attracted significant attention in medical image segmentation due to lower computational cost, such as LSegDiff \cite{vu2023lsegdiff}, SDSeg \cite{lin2024stable}.
% LDSeg \cite{zaman2024latent} 
% Diffusion-based models by Wolleb et al.~\cite{wolleb2022diffusion}, MedSegDiff~\cite{wu2024medsegdiff} and MedSegDiff-V2~\cite{wu2024medsegdiffv2} operate in pixel domain.
% Wolleb et al. \cite{wolleb2022diffusion} introduced an implicit ensemble framework using stochastic sampling from a diffusion model. MedSegDiff was introduced \cite{wu2024medsegdiff} for medical image segmentation in pixel space and later was improved in MedSegDiff-V2 \cite{wu2024medsegdiffv2}, incorporating Transformer architectures in the pixel domain. 
% To tackle computational cost at pixel space, diffusion models in the latent space have been adopted, including LDSeg \cite{zaman2024latent} and LSegDiff \cite{vu2023lsegdiff} and SDSeg \cite{lin2024stable}.
% which employ a conditional DDPM-based model in the latent space. SDSeg \cite{lin2024stable} adopts a single-step reverse process based on Stable Diffusion. 
% with the widespread success of Contrastive Language–Image Pre-training (CLIP) in understanding text prompts within natural image domains, 
In parallel, several studies have adapted Contrastive Language–Image Pre-training (CLIP)-based models to natural-language–prompted medical segmentation, such as MedCLIP-SAMv2~\cite{koleilat2024medclipsamv2}, CLIP-Driven universal model~\cite{CLIPDrivenUniversalModel}.
% and FlanS~\cite{FlanS}. 
However, all aforementioned methods require training from scratch with large-scale annotated datasets and a substantial computational cost. In contrast, our approach leverages the \emph{rich cross-modal priors encoded in a pre-trained diffusion model} to achieve effective semantic alignment between natural language prompts and medical images and their segmentation, without training on large-scale datasets. To date, the advantages of the diffusion-based model for prompt-guided segmentation remain largely underexplored.
% \paragraph{Prompt-guided Medical Image Segmentation:} 
% Early adoption of prompt-based segmentation primarily relies on spatial prompts such as point or bounding box~\cite{ma2024segment, Wu2023MedicalSA}. However, it limits its use cases, as one must first identify the object of interest beforehand. 
% Instead, natural language prompts provide a more intuitive interface for interactive segmentation. 
 
% Recently, TGEDiff~\cite{TGEDiff} explored the benefit of textual information in diffusion-based medical image segmentation, but it relies on internally derived labels, such as the number and size of target regions, from segmentation masks as auxiliary supervision, rather than utilizing user-driven prompts.
% TextDiffSeg \cite{ma2025textdiffseg} injects text-enhanced image features into a diffusion model, attaching them to the denoiser’s encoder and training with an additional segmentation loss. In contrast, we use a ControlNet-like scheme that conditions the decoder of a pretrained, frozen denoising U-Net and optimize only the diffusion loss.

% \paragraph{Low-Rank Adaptation for Medical Image Segmentation:} 
Low-Rank Adaptation (LoRA) fine-tuning has recently emerged as an efficient strategy for transferring large-scale pre-trained vision models to medical image segmentation tasks~\cite{Zhu2023MeLoLA, paranjape2025low}. 
Nevertheless, adapting diffusion-based segmentation models across modalities using LoRa fine-tuning remains unexplored.
% In the context of cross-modality adaptation, PEMMA~\cite{saadi2024pemma} introduces a LoRA-based plug-and-play framework for adapting the segmentation model trained on CT scans to positron emission tomography scans without catastrophic forgetting. 
% Recently, CtrLoRA~\cite{xu2024ctrlora} demonstrated that LoRa-based fine-tuning of ControlNet effectively adapts models to new conditions. Nevertheless, adapting diffusion-based segmentation models across modalities using LoRa fine-tuning remains unexplored.
% Motivated by the demonstrated success of LoRA-based applications and inspired by CtrLoRA~\cite{xu2024ctrlora}, ControlNet can potentially achieve few-shot segmentation through LoRA-based fine-tuning.

% Prompt-guided medical image segmentation provides a flexible and interpretable paradigm for user-driven control. MedSAM ~\cite{ma2024segment} and One-prompt~\cite{Wu2023MedicalSA} have explored the use of spatial prompts to guide segmentation toward specific target regions, demonstrating exceptional segmentation quality. Compared to spatial prompts, natural language prompts offer a more intuitive interface for interactive segmentation. With the widespread success of Contrastive Language-Image Pre-training (CLIP) in text prompt in natural images, many studies have adopted CLIP models to the medical domain for the segmentation tasks.

% These advancements highlight potential of LoRA for enabling efficient multi-modality transfer in medical image segmentation. 

%% file: section/03_method.tex
\section{Method}
\label{sec:meth}

\begin{figure*}[t]
    \centering
    \includegraphics[width=0.9\textwidth]{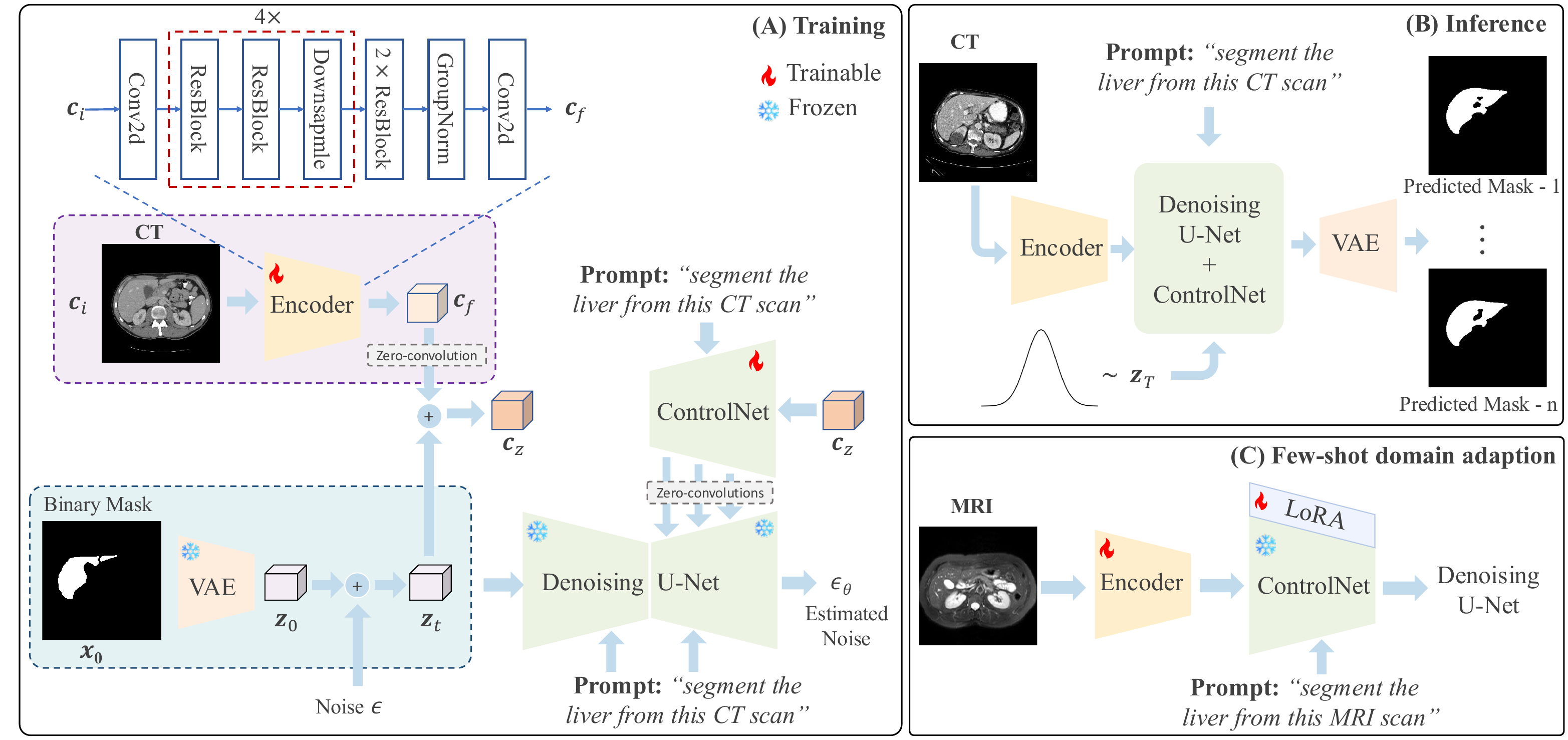}
    \caption{Overview of our proposed ProGiDiff method. We learn a control net and a custom encoder (A) to steer pre-trained Stable Diffusion to generate prompt-conditioned segmentation. Note that this formulation reduces the number of trainable parameters compared to training Stable Diffusion and the controller from scratch. During inference (B), our model is able to generate multiple candidate segmentations. We also introduce a few-shot adaptation of the learned prompt-segmentation relationship to different imaging modalities via low-rank adaptation (C).}
    \label{fig:pipeline}
\end{figure*}

In this section, we present the problem statement and describe our proposed framework for training our custom encoder and the ControlNet. Finally, we describe our strategy for cross-model domain adaptation.

\paragraph{Problem Statement:} We consider an image $\boldsymbol{c}_i\in\mathbb{R}^{H\times W}$. Our objective is to generate segmentation masks $\boldsymbol{x}_l\in\{0,1\}^{H\times W}$ for class $l\in \{1,2,\cdots L\}$ for $L$ number of classes conditioned on a natural language prompt $y_l$, which is derived based on the class. We aim to learn a diffusion-based model $f_\theta$ to produce $\boldsymbol{x}_l=f_\theta(\boldsymbol{c}_i,y_l)$. Specifically, we aim not to learn $f_\theta$ entirely from scratch, but to steer a pretrained diffusion model for image generation to produce segmentation masks conditioned on an image and prompt.

\subsection{ProGiDiff: Steering Diffusion for Segmentation}
The entire workflow of ProGiDiff is illustrated in Figure~\ref {fig:pipeline}. 
% As mentioned earlier, our objective is to generate segmentation from medical images conditioned on a natural language prompt for that class. 
% We explore novel research directions as posed in Section \ref{sec:intro}, Q1, and Q2.
% on the possibility of steering a pretrained diffusion model and leveraging its text condition mechanism in doing so.
We base our framework on the SOTA image generation method Stable Diffusion (SD). 
% It comprises of a pre-trained variational autoencoder (VAE) for bidirectional mapping between the image and latent space, a denoising U-Net, and a prompt-conditioning mechanism to guide generation. 
However, for us, instead of an image $I$, the input is a binary organ mask $\boldsymbol{x_0}\in \mathbb{R}^{H \times W \times 3}$ (where $H, W$ are the height and width, respectively). $\boldsymbol{x_0}$ is mapped to the latent space by the VAE encoder~\(\mathcal{E}\), $\boldsymbol{z}_0 = \mathcal{E}(x_0), \boldsymbol{z}_0 \in \mathbb{R}^{h \times w \times c}$, (where $h, w, c$ are the height, width, and channel dimensions in the latent space, respectively). The corresponding pre-processed CT image $\boldsymbol{c}$ serves as the conditioning input to provide anatomical context for the segmentation task. Note that for the conditioning, our method does a \emph{switcheroo} of the position of image and segmentation for the original ControlNet setup. While the original ControlNet aims to generate a \emph{realistic image conditioned on segmentation}, we aim to generate \emph{faithful segmentation conditioned on images}. 
% To achieve this, we employ a custom encoder to extract semantic information from medical images, which is then used to guide the denoising of the segmentation mask. 
The conditioning vector $\boldsymbol{c_f}$ is passed through zero-convolution blocks and added to the noisy mask latent $\boldsymbol{z_t}$, before being processed by the ControlNet layers. The outputs of the ControlNet blocks are subsequently integrated into the decoder stages of the denoising UNet via an additional set of zero-convolution layers.

For the text-conditioning, we rely on a simple prompt structure, e.g., \textit{"Segment the Liver in this CT scan"}. 
% We observe that while such a strategy is simple, it introduces prompt similarity between different organs based on the CLIP embeddings. 
To mitigate prompt similarity between different organs based on the CLIP embeddings, we use synonyms of the segmentation task, e.g., "Highlight" or "Show." Text prompts are randomly selected from three predefined variants during training. 
% The pre-trained CLIP model from SD encodes the selected prompt and serves as a conditioning input to both the Denoising UNet and ControlNet. 
% guiding the model toward segmenting the specified organ. 
During inference, we observe little reliance on the task-specific keywords, and its critical dependency on the organ-specific keyword such as "Liver", hence we use only one variant.

% During training, the entire SD model is kept frozen, while only the ControlNet and the custom conditioning encoder are trained. We adopt the same loss function as SD.
% SDSeg~\cite{SDSeg}. Specifically, LDM loss in Eq.~\ref{eq:LDM_loss} is adopted as part of the training objective $\mathcal{L}_{\text{noise}}$, but we replace the standard L2 loss with an L1 norm. To enhance training supervision, a latent reconstruction loss $\mathcal{L}_\text{latent}$ is introduced to minimize the gap between $\tilde{z_0}$ and the ground truth latent $z_0 = \mathcal{E}(\boldsymbol{x})$. The predicted $\tilde{z_0}$ is computed using the estimated noise, following Eq.~\ref{eq:noisy_latent}. The overall training objective can be expressed as follows:
% \begin{equation}
% \mathcal{L} = \mathcal{L}_\text{noise} + \lambda \mathcal{L}_\text{latent}, \quad \text{where} \quad \mathcal{L}_\text{latent} = \mathcal{L}(\tilde{z}_0, z_0)
% \end{equation}
% where $\mathcal{L}_\text{latent}$ is also implemented as the mean absolute error. $\lambda$ is a weighting factor (set to 1 in our implementation). 
% \begin{equation}
% \mathcal{L}_{\text{noise}} := \mathbb{E}_{z_0, \epsilon, t} \left[ \left\| \epsilon - \epsilon_\theta(z_t, t, \tau_\phi(y)) \right\|_1 \right]
% \label{eq:L1_loss}
% \end{equation}
\subsection{Conditioning Image Encoder}

Since ControlNet is applied in a reverse manner, we replace the original shallow CNN encoder with a deeper architecture to better capture rich features from the CT/MR scan. The encoder structure follows the same design as the VAE encoder used in SD, with modifications to the channel progression. Instead of starting from 128 channels as in the original VAE encoder, our version begins with 16 channels and progressively increases the channels to 32, 64, 128, and 256. The detailed architecture is illustrated in Fig.~\ref{fig:pipeline}(A). 
% The proposed encoder contains approximately 6.9 million trainable parameters. 
% % compared to 35 million in the original VAE encoder. 
% While this represents a moderate increase over the 1.09 million parameters in the default encoder in the original ControlNet, the custom encoder significantly enhances the extraction of meaningful features from medical images, resulting in a more stable segmentation process. Importantly, this enhancement does not introduce a noticeable increase in training or inference time.

\subsection{LoRA-based Fine-tuning for ControlNet}
% Here we explore our third research question %posed in \ref{sec:intro},
% on the transferability of learned prompt-segmentation semantic relationships from CT to MRI modality. 
MR and CT images differ significantly in terms of appearance, contrast, and organ boundary characteristics. As a result, a ControlNet model trained solely on CT images lacks the ability to perform effective cross-modality segmentation when applied to MR images in a zero-shot manner. 
% To efficiently adapt the pre-trained CT-based ControlNet for cross-modality medical image segmentation, 
To this end, we adopt a LoRA few-shot domain adaptation strategy. In practice, we insert LoRA modules into all linear components of the ControlNet architecture. During fine-tuning, the added LoRA parameters, the custom encoder for the conditioning image, normalization layers, and all zero convolutional layers are updated, while the remaining model weights remain frozen. 
% It is achieved with only 10\% of the parameters of the CT-based ControlNet and with a significantly smaller number of training samples.
The training objective remains identical to SD. 
% It is achieved with a significantly smaller number of training samples. 
% This few-shot finetuning setup enables efficient domain adaptation with limited training data.
% including the projection layers for Query, Key, Value, and output tensors within each attention block, the two linear layers in each transformer's Feed-Forward Network, and the linear projections for processing conditional control inputs, such as time-step embedding inputs. 

%

% \subsection{Loss Function}
% This part can be reused from my thesis.

%% file: section/04_exp.tex
\section{Experiments}
\label{sec:exp}

\begin{table*}[t]
    \centering
    \setlength{\tabcolsep}{3pt}
    \begin{tabularx}{\linewidth}{lccccccccccc}
        \toprule
        \textbf{Model}   & Generative & Prompt-Guided & \textbf{Liver} & \textbf{Spleen}  & \textbf{L.kidney} & \textbf{R.kidney} & \textbf{Stomach} & \textbf{Pancreas} & \textbf{Dice$\uparrow$} & \textbf{HD95$\downarrow$} \\
        \midrule
        % nnUNet(3D)      & \textbf{97.16} & \textbf{91.95} & \textbf{87.29} & \textbf{88.05} & \textbf{89.58}  & \textbf{82.90} & \textbf{89.49}\\
        TransUNet (3D)  & \xmark & \xmark & 93.40 & 83.13 & \textbf{85.44} & \textbf{86.95} & 82.37 & \textbf{80.33} & 85.27 & 16.36 \\
        nnUNet (2D)      & \xmark & \xmark & \textbf{95.63} & 88.87 & 85.34 & 81.99 & 79.29 & 70.89 & 83.67 & 16.39\\
        \midrule
        Diff-UNet (3D)   & \cmark & \xmark & 93.34 & 87.60 & 77.75 & 84.84 & 75.91 & 71.09 & 81.76 & 11.36 \\
         \midrule
        MedCLIP-SAMv2   & \xmark  & \cmark & 24.73    &  21.58      &   6.82   &   4.55    &   24.32   &   6.82    &  14.80  & 92.29\\
        \midrule
        Ours            & \cmark  & \cmark & 94.56 & 88.63 & 79.29 & 80.41 & 58.89 & 48.39 & 75.03 & 19.06\\
        Ours (Oracle@50)  & \cmark & \cmark & 95.52 & \textbf{93.71} & 84.25 & 85.05 & \textbf{85.22} & 71.13 & \textbf{85.81} & \textbf{3.89}\\

        \bottomrule
    \end{tabularx}

    \caption{Quantitative comparison of abdominal multi-organ segmentation on the BTCV dataset between ProGiDiff and SOTA 2D/3D methods.
    Results are reported as Dice Similarity Coefficients (\%) and 95\% Hausdorff Distance (mm). 
    Best score for each category is shown in \textbf{bold}.}
    \label{tab:baseline_comparison}
\end{table*}

\subsection{Dataset:}We use the BTCV dataset~\cite{btcv} for organ segmentation on CT images and the CHAOS dataset~\cite{chaos} for few-shot cross-modality adaptation on MR images. 
Following the protocol of TransUNet~\cite{chen2021transunet}, we divide the dataset into 18 training and 12 testing cases. For MRI experiments, we use 19 annotated T2-weighted abdominal scans from the CHAOS dataset, excluding one subject due to noticeable intensity inhomogeneity in the liver and spleen regions and one subject in ksidneys.
The image preprocessing and experiment implementation details are described in the repository.

\subsection{Baselines \& Metric:} We compare our method with multiple strong deterministic baselines in 2D~\cite{isensee2021nnu} and 3D~\cite{3Dtransunet}. 
% including 2D nnUnet~\cite{isensee2021nnu} and 3D TransUNet~\cite{3Dtransunet}. 
For the diffusion-based baseline, we compared it against the recent 3D Diff-Unet~\cite{xing2023diff}. 
% It handles multi-class segmentation in 3D.
% Note that this is a strong representative of the diffusion-based model categories, as it handles multi-class segmentation in 3D. 
Additionally, we compare against MedCLIP-SAMv2~\cite{koleilat2024medclipsamv2}, a prompt-guided 2D model for medical image segmentation. Note that our proposed ProGiDiff is a 2D model. We use the Dice and 95\% Hausdorff Distance metric for the segmentation evaluation. Further, as an explorative study, we investigate the possible upper bound of our method in the presence of an \emph{oracle}, fully leveraging multiple proposals from our model. We report the evaluation in Table \ref{tab:baseline_comparison}. Note that although Diff-UNet is a generative model, its predictions are very similar to each other with a minute difference, which did not improve in the presence of an oracle.

\begin{table}[!t]
    \centering
    \begin{tabularx}{\linewidth}{lcccc}
        \toprule
        \textbf{Training Method} & \textbf{Liver} & \textbf{Spleen} & \textbf{L.Kid} & \textbf{R.Kid} \\
        % & \textbf{Average} & \textbf{Params(M)}\\
        \midrule
        Ours from Scratch       & 87.11 & 61.59 & 26.46 & 71.27 \\
        % & 61.61 & 367.13 \\
        Diff-UNet (r=64) & 84.30 & 82.45 & \textbf{76.28} & 81.34 \\
        % & 80.56 & 2.79 \\
        %  Diff-UNet (r=128) &  &  &  &    \\
        Ours (r=64)   & \textbf{88.22} & \textbf{89.05} & 70.68 & \textbf{87.55} \\
        % & \textbf{81.25} & \textbf{31.18} \\
        % Ours (r=128)  & \textbf{89.20} & 85.42 & 67.26 & \textbf{78.10} \\
        % & 81.59 & 43.89  \\
        \bottomrule
    \end{tabularx}

    \caption{Few-shot cross-modality adaptation on the CHAOS MRI dataset, comparing LoRA fine-tuning of ControlNet vs. Diff-UNet vs. training from scratch. 
    Dice (\%) and trainable parameters (M) are reported. 
    Best results are in \textbf{bold}.}
    
    \label{tab:MR_few_shot}
\end{table}

\subsection{Main Results:}
% Table 1: comparing with SOTA on CT abdominal multi-organ segmentation

We evaluate our method on abdominal multi-organ segmentation using the BTCV dataset, with results shown in Table~\ref{tab:baseline_comparison}. 
The table highlights two key advantages of our approach: the generative capability and prompt-guided flexibility, which are unique features not simultaneously present in existing SOTA methods.
% Despite being a 2D approach, our base model achieves competitive performance compared to both 2D and 3D methods, particularly on large organs such as the liver and spleen. 
% Performance on smaller or more complex structures like the pancreas and stomach is initially lower, likely due to likely due to reduced feature discriminability in these ambiguous regions, resulting in weaker semantic representations. Supporting evidence is provided in the Supplementary Material.
By leveraging our model’s generative nature through Oracle@50, which selects the best segmentation among multiple candidates, 
% we achieve a substantial performance boost compare to our base model. 
% (nearly 10 percentage points on average from 75.05\% to 84.31\%).
% Specifically, 
our method attains top performance on the spleen, stomach and achieves the best averaged Dice and HD95 scores, outperforming both the SOTA baselines and our base model. 
This dramatic improvement underscores the strength of diffusion-based generation in producing multiple plausible segmentations and providing clinicians with a range of valid hypotheses for decision support, i.e., a capability not found in traditional deterministic models.
Performance on smaller structures like the pancreas is initially lower, likely due to reduced feature discriminability in these ambiguous regions, resulting in weaker semantic representations. 
% Supporting evidence is provided in the Supplementary Material.
The notably worse performance of MedCLIP-SAMv2 can be attributed to the absence of domain adaptation for abdominal organs and its evaluation in a few-shot setting.
% both of which likely limit its ability to segment fine-grained anatomical structures in CT.

Overall, these results are particularly encouraging, as our 2D approach demonstrates strong performance relative to established 3D methods, highlighting the potential of leveraging language-guided diffusion for medical segmentation. 

Additionally, an ablation study demonstrates that the customized condition image encoder outperforms the original on five of six organs, yields larger gains on challenging organs (pancreas, stomach), and increases the average Dice by 6.09\% (from 70.72\% to 75.03\%).
% Visual representations are provided in the Supplementary Material.

% \begin{figure}[!t]
%     \centering
%     \includegraphics[width=1.0\linewidth]{figures/mri_images_1.png}
%     \includegraphics[width=1.0\linewidth, trim={0 0 0 20},clip]{figures/mri_images_2.png}
%     % \includegraphics[width=1.0\linewidth, trim={0 0 0 20},clip]{figures/mri_images_3.png}
%     % \includegraphics[width=1.0\linewidth, trim={0 0 0 20},clip]{figures/mri_images_4.png}
%     \caption{Visual comparison of LoRA fine-tuning and from-scratch training on the CHAOS MRI dataset.}
%     \label{fig:mri_visual}
% \end{figure}

% Table 2: MRI abdominal multi-organ segmentation
\begin{figure}[!t]
    \centering
    \tiny{
    \hfill Image \hspace{3.5em} nnUnet  \hspace{2.5em} TransUnet \hspace{3.0em} Diff-Unet \hspace{3.0em} Ours \hspace{1.5em} Ours(Oracle@50) \hspace{1.5em} Ground Truth} \hfill
    \includegraphics[width=1.0\linewidth, trim={0 0 0 20},clip]{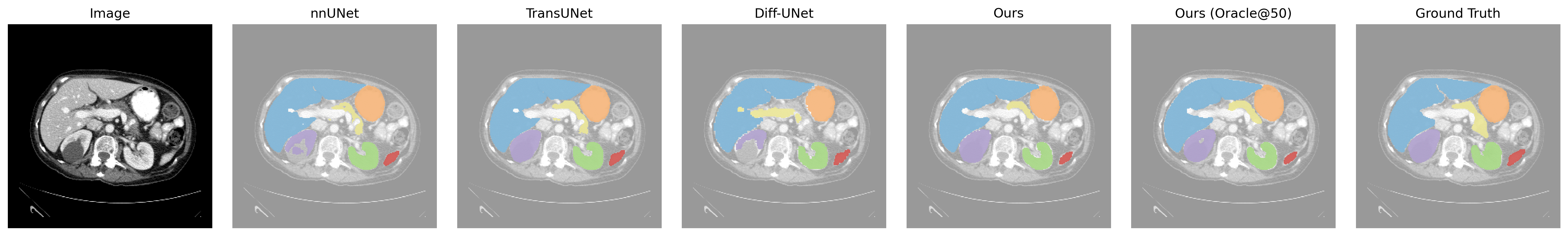}
    \includegraphics[width=1.0\linewidth, trim={0 0 0 20},clip]{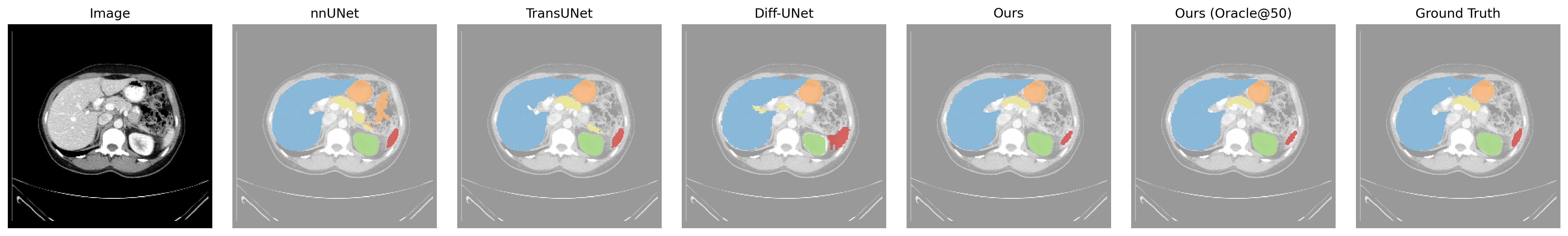}
    \caption{Qualitative comparison of our method with SOTA segmentation models on the BTCV Dataset.}
    \label{fig:CT_visual}
\end{figure}
\subsection{Few-shot Adaptation on MRI:}

Table~\ref{tab:MR_few_shot} presents cross-modality adaptation results to MR images in a few-shot setting. 
% comparing LoRA fine-tuning with Diff-UNet and with our model trained from scratch.
% Training ControlNet from scratch yields moderate performance despite using 367.13M parameters. In contrast, LoRA fine-tuning of ControlNet achieves substantially better results using only 31.18M parameters, with notable gains across all organs. 
LoRA fine-tuning of ControlNet achieves substantially better results with only 31.18M parameters, showing notable gains across all organs compared with training from scratch with 367.13M parameters.
And its average Dice (83.88\%) is higher than that of Diff-UNet (81.09\%). Nevertheless, our model is more sensitive to intensity inhomogeneity, which can result in outliers.
% particularly the spleen (85.84\%) and right kidney (73.85\%).
% Interestingly, increasing the LoRA rank to 128 results in only marginal improvements for the liver and left kidney, while using more parameters and slightly lowering the overall average compared to rank 64.
% This indicates that a moderate rank offers a better trade-off between parameter efficiency and segmentation accuracy.
These results demonstrate that LoRA can effectively adapt our pre-trained model to a new imaging modality using substantially fewer parameters.
% achieving higher accuracy with only about 10\% of the parameters required for full training.
Such efficiency is especially valuable in medical image analysis, where labeled data is often limited.

\subsection{Simulated Expert-in-Loop and Ensemble:}

While the best result from our method is obtained via an oracle, i.e., selecting the best proposal out of $k$ predictions, this upper bound is not a practical solution.
% which serves as an upper bound for the expected performance, it is not a practical solution. 
% To explore this further, we simulate a human expert-in-the-loop scenario by not selecting the best option every time, but rather a random one from the top 3 proposals, modeling inter-rater variability, referred to as RandTop-3. 

\begin{figure}[!t]
    \centering
    \includegraphics[width=0.8\linewidth]{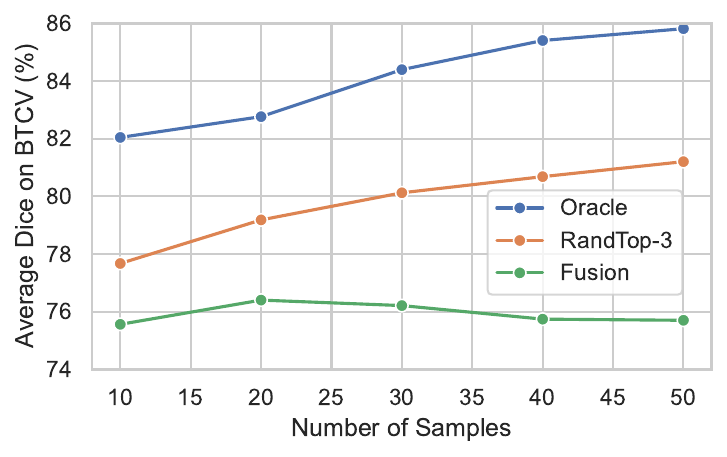}
    \caption{Performance comparison of a simulated expert-in-loop (RandTop-3) with respect to oracle and fusion for varying numbers of generated segmentations.}
    \label{fig:diff_samples}
\end{figure}

To explore this further, we simulate a human-in-the-loop setting by randomly selecting one of the top-3 proposals (RandTop-3) to model inter-rater variability.
We show in Fig. \ref{fig:diff_samples} how this simulated expert behaves with the number of proposals in comparison to the oracle. 
% We observe a consistent performance boost with an increase in the number of proposals. We also report our ensemble over an increasing number of samples and observe that it does not offer a significant performance boost. We attribute this to an increase in uncertainty due to the higher number of proposals.
We observe consistent gains as the number of proposals increases, but ensembling more samples yields little additional benefit, likely due to increased uncertainty.

